\documentclass[11pt, a4paper, onecolumn, copyright, numbering]{deepmind}

\usepackage[authoryear, sort&compress, round]{natbib} 

\graphicspath{{figures/}}

\title{Rapid Object Annotation}

\correspondingauthor{mdenil@google.com}

\renewcommand{\today}{2020-08-27}

\author[1]{Misha Denil}

\makeatletter
\AtBeginDocument{
  \hypersetup{
    pdftitle = {\@title},
    pdfauthor = {\@author}
  }
}
\makeatother

\begin{abstract}
In this report we consider the problem of rapidly annotating a video with bounding boxes for a novel object.
We describe a UI and associated workflow designed to make this process fast for an arbitrary novel target.
\end{abstract}

\begin{document}
\maketitle

\tableofcontents

\clearpage

\section{Introduction}

We envision the following workflow to train a detector for a novel object:
\begin{enumerate}
\item Record a video of the target object from many angles.
\item Annotate the video frames to identify the target object.
\item Use the annotations to fine tune a detection model.
\end{enumerate}
In terms of human time and effort the second task is by far the most expensive.
Recording a video is easily done using a cell phone, and fine tuning a model on a new data set is a standard task in machine learning, and the setup time can be amortized by standardizing data formats.
The remaining challenge is to make annotation of the collected video easy and fast.
We focus on optimizing the use of human time in this process.

We design an annotation tool that provides assistance to the operator by taking advantage of a pretrained objectness prior to identify the extent of objects and to propagate labels between frames.
We evaluate the tool by annotating several videos and measuring the number of high quality bounding boxes obtained per second of human annotation time.

\section{Annotation tool}

\begin{figure}
    \centering
    \includegraphics[width=\linewidth]{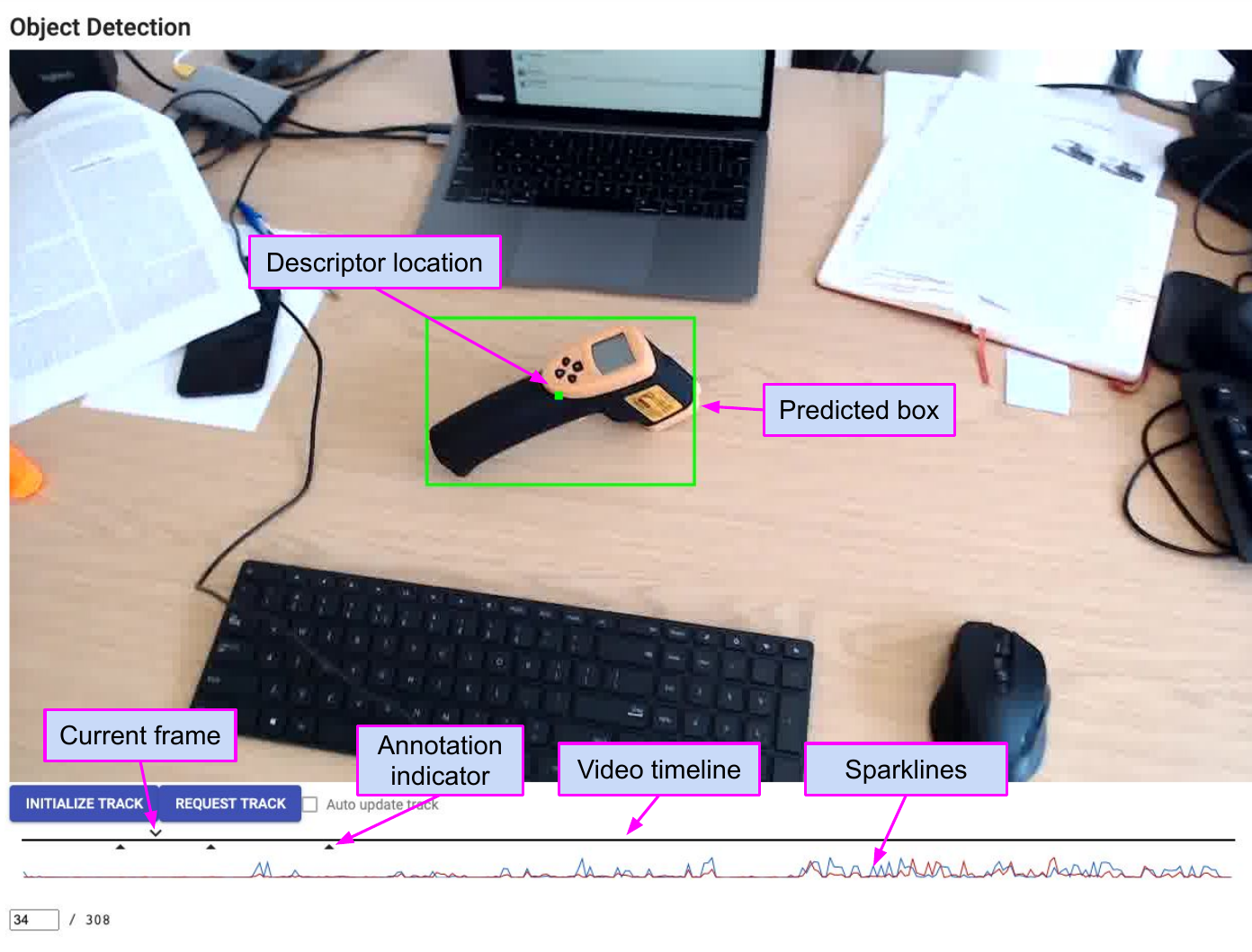}
    \caption{Example of the annotation UI.}
    \label{fig:ui-example}
\end{figure}

The annotation UI is shown in Figure~\ref{fig:ui-example}.
The UI is composed of three main components:
\begin{itemize}
\item The \textbf{viewport} shows a single video frame, as well as any annotations and predictions associated with that frame.
\item The \textbf{timeline} situates the frame in time.  It shows an indicator for the current frame shown in the viewport, as well as markers to indicate frames for which there are annotations.
\item The \textbf{sparklines} show indicators about the relationship of adjacent frames of the track to each other.  This is useful for identifying when object identity or position may have been lost.
\end{itemize}

The rest of this section describes the tool, but the easiest way to get a feel how it works is to \href{https://www.youtube.com/watch?v=Rrzayts68zY}{watch a video of it being used}.

\subsection{Viewport}

The viewport displays a frame, as well as descriptor locations and boxes.
Each frame can have one or zero annotated locations associated with it.
Annotated locations (and their associated boxes) are shown in a different color than predictions so that they can be easily distinguished.

\begin{figure}
    \centering
    \includegraphics[width=0.5\linewidth]{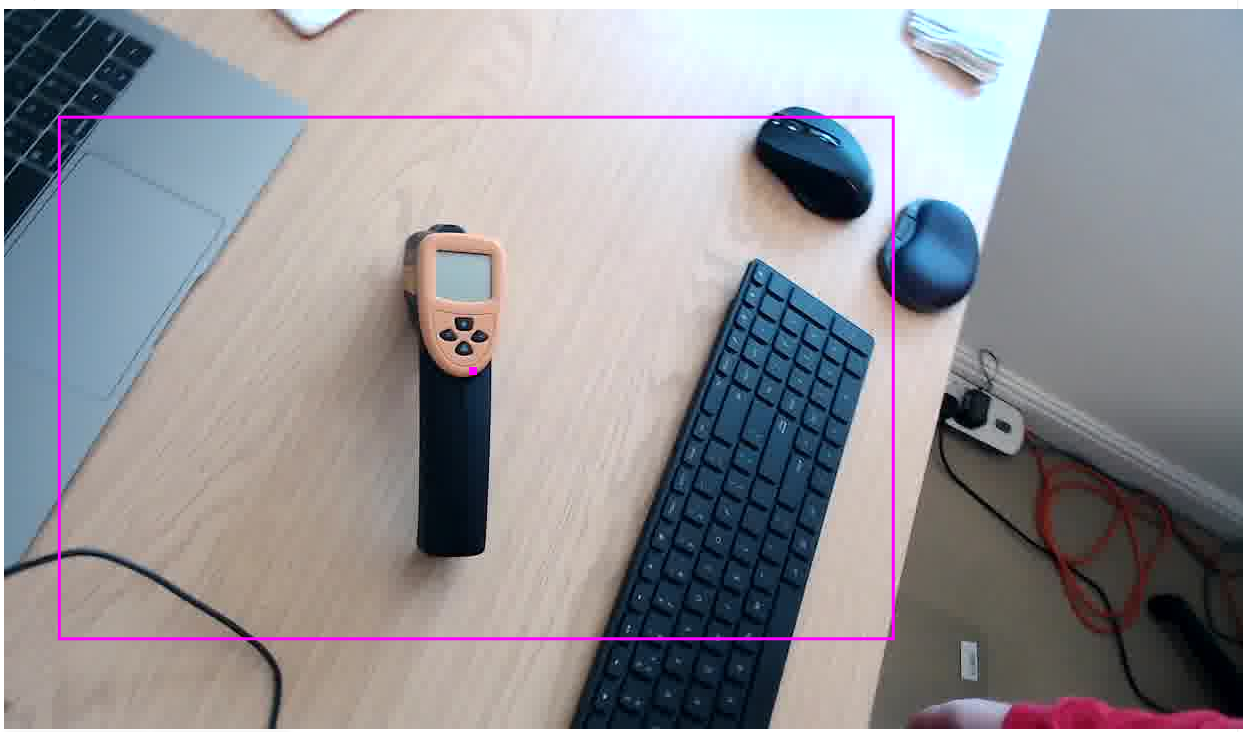}
    \caption{Example of why it is useful to show the bounding boxes. The descriptor location in this example seems reasonable for the target object, but the predicted bounding box shows that the descriptor content is not.}
    \label{fig:poor-grounding}
\end{figure}

In Figure~\ref{fig:poor-grounding} we show why it is useful to display more than just the descriptor location in the viewport.
In this case the descriptor location (at the bottom of the yellow face of the thermometer) appears reasonable for the target object, but the bounding box produced by the descriptor at this location shows that the discriptor content is more likely associated with the table than the target object.

\subsection{Timeline}

The timeline allows navigation through the video and shows indicators for which frames have been annotated.
A cursor above the tineline shows the location of the current frame in the video, and clicking on the timeline moves the current frame to the location of the click.
Indicators under the timeline show the locations of annotated frames.

\subsection{Autotrack}

Autotracking allows a set of sparsely annotated frames in a video to be extended to a label for every frame.
The mechanism for propagating labels is described in Section~\ref{sec:autotrack}.

\subsection{Sparklines}

Sparklines appear under the timeline and are aligned with it.
They show information about how the track produced by autotracking changes from frame to frame.
Specifically there are two lines, blue and red, which show the change in descriptor location and change in box area from frame to frame, respectively.
These lines can be used to quickly identify potential parts of the video where automatic tracking has failed, allowing the operator to focus attention on inspecting these areas for further annotation.

\subsection{Smartjump}

Smartjump is a feature that allows automatically jumping to the frame where the current track shifts the most.
A large jump in track location often indicates loss of tracking, making this a good frame to annotate.

\subsection{Extreme clicking}

Extreme clicking \citep{papadopoulos2017extreme} creates axis aligned bounding boxes from a set of ``extreme'' points.
To create a bounding box the operator clicks four times, once each at the two horizontal and two verical extremes of the target object.
The resulting annotation is the smallest axis aligned bounding box that contains the selected points.
We use extreme clicking in this report as the baseline exisiing method for bounding box annotation to compare against, and also for the creation of ground truth data to evaluate the lables produced by our tool.

\section{Experiments}

\subsection{Target objects}

\begin{figure}[h]
    \centering
    \includegraphics[width=0.8\linewidth]{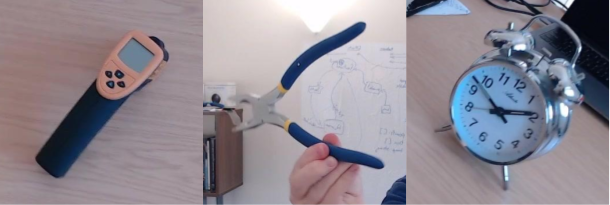}
    \caption{The three target objects we consider in this report.  From left to right they are an infra-red thermometer, a pair of pliers, and a clock.}
    \label{fig:object-examples}
\end{figure}

We use three semi-arbitrarily chosen objects as the targets for the experiments.
An example of each object is shown in Figure~\ref{fig:object-examples}.

The infra-red thermometer was chosen because it is a sort of familiar looking object that does not appear in the label set of MSCOCO, and is also quite unlikely to have appeared unlabeled in MSCOCO incidentally.
The pliers were chosen because they are articulated and also highly non-convex, which potentially makes labeling harder.
The alarm clock was chosen because I happen to have one in my apartment.

\subsection{Videos}

For each target object we collect two videos.
The videos are trimmed to 30 seconds and resampled at 10 fps so that each one provides exactly 300 frames.

For each object the first video is taken with the camera held in hand and the object sitting on a desk.
The camera is moved around by the operator to collect different views of the object, and occasionally the operator reaches into the frame to reorient the object (for example by flipping it over).

The second video of each object is captured by a stationary mounted camera.
The operator stands in the camera frame and holds the object in view, moving it around so that it is captured from multiple viewpoints.

\subsection{Annotation styles}

We use the tool to annotate several videos with different sets of features enabled.
In the following we distinguish between a \emph{label} and an \emph{annotation}.
A label is a point and a bounding box (in some configurations we record only a point) associated with a frame.
An annotation is a label produced by human input.
We collect labels for all frames in all videos.
Depending on the annotation style this may be a result of much fewer annotations.

\paragraph{XClick}
Frames are presented in a random order, and bounding boxes and center points are annotated in each frame using extreme clicking \citep{papadopoulos2017extreme}.  Data produced by extremely clicking is considered ``ground truth'' when such a thing is appropriate.  In this configuration the video timeline and sparklines are hidden.

\paragraph{Click}
Frames are presented in a random order and the operator is asked to click on the center of the target object in each frame.
The selected location is shown in the viewport, but no additional feedback is given.
The operator can change their selected point by clicking again.
In this configuration the video timeline and sparklines are hidden.

\paragraph{Boxes}
Frames are presented in a random order and the operator is asked to click on the center of the target object in each frame.
The selected location in the UI along with a predicted bounding box for the selected point.
The operator can change their selected point by clicking again.
In this configuration the video timeline and sparklines are hidden.

\paragraph{Autotrack}
Frames are presented using the video timeline.
The operator can navigate through frames using the standard timeline controls.
The operator can click a point in the image to annotate a point in the current frame, and can also press \texttt{r} to request the system fill in a track for unannotated frames.
Predicted boxes are shown for the track but not for the annotated points.
In this configuration the sparklines are hidden.

\paragraph{Autotrack-boxes}
The same configuration as \textbf{Autotrack} but predicted boxes are also shown for the annotated points.

\paragraph{Autotrack-boxes-sparklines}
The same configuration as \textbf{Autotrack-boxes} but the sparklines are no longer hidden.

\paragraph{Autotrack-boxes-sparklines-smartjump}
The same configuration as \textbf{Autotrack-boxes-sparklines} but jumping nativgates to the frame with the largest jump in track position, instead of to a random unannotated frame.

\section{Results}

\subsection{Annotation time}

\begin{figure}
    \centering
    \includegraphics[width=0.45\linewidth]{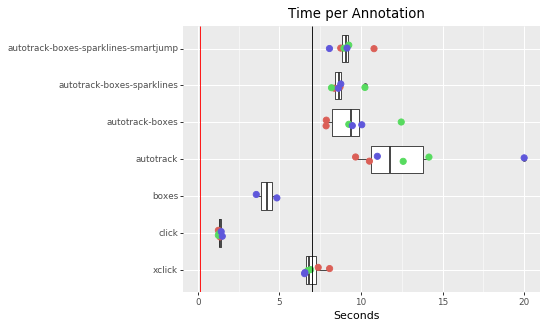}
    \quad
    \includegraphics[width=0.45\linewidth]{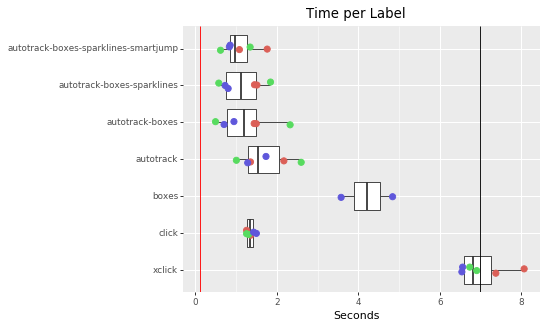}
    \caption{Annotation times under different tools settings, aggregated across videos and objects.  The red vertical lines indicate ``real time.''  The black vertical lines show the time per box reported elsewhere for xclick.}
    \label{fig:annotation-time}
\end{figure}

\begin{figure}
    \centering
    \begin{minipage}{0.45\textwidth}
    \includegraphics[width=\linewidth]{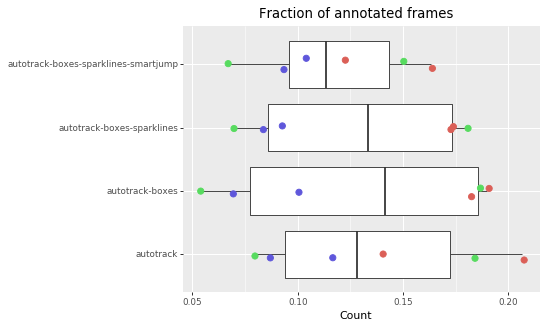}
    \end{minipage}
    \hfill
    \begin{minipage}{0.5\textwidth}
    \begin{tabular}{l|r}
\textbf{Setting} & \textbf{Annotation Time} \\
\hline
xclick & \texttt{3:30:45} \\
autotrack & \texttt{0:50:44} \\
boxes & *\texttt{0:42:03} \\
click & \texttt{0:40:31} \\
autotrack-boxes & \texttt{0:37:02} \\
autotrack-boxes-spark & \texttt{0:34:33} \\
autotrack-boxes-spark-jmp & \texttt{0:32:27} \\
\hline
Total & \texttt{7:28:05}
\end{tabular}
\end{minipage}
    
    \caption{\textbf{Left:} Fraction of annotated frames for each method where annotating less than 100\% of frames is possible. \textbf{Right:} Total time spent annotating in the different styles.  Each annotation style counts time to annotate 6 videos, except for ``boxes'' counts time to annotate 2 videos.  If we had annotated all 6 videos using boxes we would expect it to have taken \texttt{2:32:12}.}
    \label{fig:fraction-of-annotated-frames}
\end{figure}

The distributions of annotation times for each configuration are shown in Figure~\ref{fig:annotation-time}.
We show both the time per annotation and time per label, which are in general not the same.
For \textbf{xclick}, \textbf{click} and \textbf{boxes} every frame is annotated so the time per annotation is equal to the time per label.
For the other settings annotation proceeds until the operator is satisfied that every frame has a label, which leads to different numbers of annotations for each video.
The fraction of annotated frames for each run is shown in Figure~\ref{fig:fraction-of-annotated-frames}.

We can compare ourselves to numbers reported elsewhere on xclick for bounding box annotation.
Our average xclick times are slightly faster than they report, likely because our tasks are slightly simpler (with exactly one object per frame and no need to indicate object class).

All annotation methods were performed on all videos, with the exception of \textbf{boxes}, which was only done for the two thermometer videos.
The decision to not include boxes was made because it is quite time consuming to annotate every frame and it was clear early on that this approach fairs poorly in the time per label metric.
The following table shows time spent annotating videos in each configuration.

\subsection{Label quality}

In addition to timing statistics we also examine the quality of the labels produced by our tool.
In order to do this we treat the bounding boxes from xclick as ground truth bounding boxes (and the bounding box centers as ground truth points).
XClick bounding boxes are most likely to be reliable, given that every frame is annotated by an operator (i.e.\ all labels are annotations) and the annotation method has been shown already to be highly reliable.

\begin{figure}
    \centering
    \includegraphics[width=0.45\linewidth]{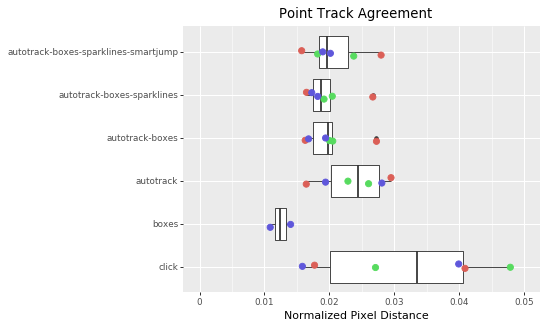}
    \quad
    \includegraphics[width=0.45\linewidth]{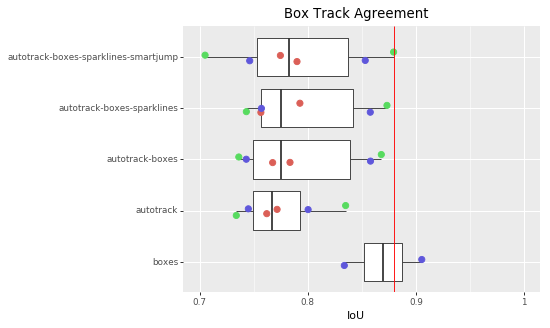}
    \caption{Agreement between xclick annotations and different annotation configurations using our tool. \textbf{Left:} Discrepancy in the point tracks (center points of the labeled bounding boxes) measured in normalized pixel coordinates (lower is better; 0 is perfect match).  \textbf{Right:} Discrepancy in the box tracks, measured in units of intersection over union (higher is better; 1 is perfect match).}
    \label{fig:track-agreement}
\end{figure}

In Figure~\ref{fig:track-agreement} (left) we show the discrepancy between ground truth point track and point tracks produced by our tool.
Discrepancy is measured in normalized pixel distance, that is a metric where the image is covered by the coordinates $[0, 1] \times [0, 1]$, making the units insensitive to the image size.

In Figure~\ref{fig:track-agreement} (right) we show the agreement between the ground truth box track and the box tracks produced by our tool.
Note that ``click'' does not appear in the box track agreement figure because that annotation style does not produce a box track.
Agreement is measured as the intersection over union of the label and ground truth box, which gives unitless values independent of the image size.
We also mark the value $0.88$ on the figure, which was shown by \cite{papadopoulos2017extreme} to be the approximate limit of average inter-annotator agreement on Pascal VOC data.
We should not expect to achieve average IoU values higher than this threshold.

\begin{figure}
    \centering
    \includegraphics[width=0.45\linewidth]{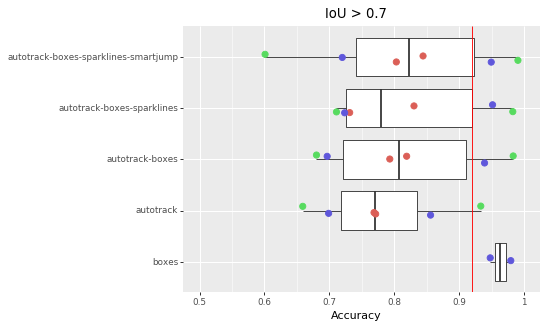}
    \quad
    \includegraphics[width=0.45\linewidth]{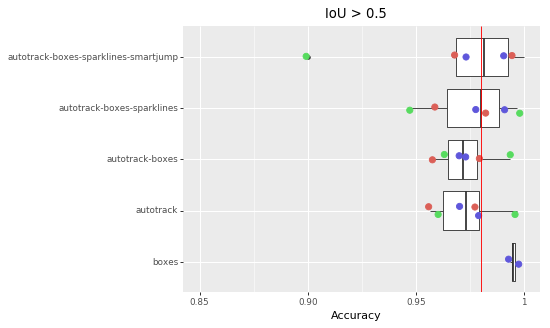}
    \caption{Accuracy of the different annotation methods when evaluated as a detector with xclick boxes as ground truth.  We do not compute average precision scores since there is always exactly one ``detection'' for each ground truth instance.  Instead we compute accuracy of the labels by thresholding IoU with the xclick boxes to determine if a label is correct.  Red lines shown the value of the corresponding metric for extreme clicking as evaluated on Pascal VOC (see~\cite{papadopoulos2017extreme}, Table 1; we show the stronger value from Pascal 2007 and Pascal 2012 in both cases).  Note that the x-axis has a different scale between the two plots.}
    \label{fig:iou-accuracy}
\end{figure}

The standard metrics for evaluating detectors involve computing a precision-recall curve by varying a detection threshold and then computing the area under that curve (with some embellishments, so it's not precisely the AUC\footnote{\url{https://medium.com/@jonathan_hui/map-mean-average-precision-for-object-detection-45c121a31173}}).
Since we are evaluating annotations rather than detections there is no detection threshold upon which to base an AUC curve.
Instead we use a threshold on IoU to determine if each bounding box is correct, and compute the accuracy of the box tracks produced by each annotation method.  Following \cite{papadopoulos2017extreme} we evaluate using IoU thresholds of 0.7 and 0.5, and show the resulting accuracies in Figure~\ref{fig:iou-accuracy}.

\begin{figure}
    \centering
    \includegraphics[width=0.45\linewidth]{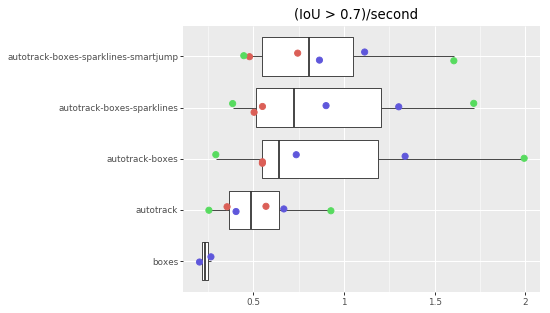}
    \caption{Number of high quality bounding box labels per second of annotation time.  The x-axis is in units of high quality boxes per second.}
    \label{fig:iou-per-second}
\end{figure}

Finally we combine the time and accuracy metrics to measure the proportion of high quality bounding boxes per unit time.
We arbitrarily pick an IoU threshold of 0.7 to be a ``high quality'' bounding box and in Figure~\ref{fig:iou-per-second} we show the number of such bounding boxes produced per second using the different annotation styles.

It seems reasonable to expect about 0.75 high quality bounding boxes per second using our annotation method, which should be contrast with 0.14 bounding boxes per second using XClick.
This corresponds to a 5.3x increase in annotation speed beyond XClick, and is likely to be a slight over estimate because it assumes that every XClick box is high quality.

\section{Methods}

\subsection{CenterNet}

CenterNet~\citep{zhou2019objects} is a model for ``Object detection, 3D detection and pose estimation.''
It was later extended by the same authors to a tracking model~\citep{zhou2020tracking}.
The tracking model is not discussed in this report, we do something much more naive for tracking here.
Incorporating ideas from the centernet tracking paper is probably a good idea for future work.

\begin{figure}
    \centering
    \includegraphics[width=0.5\linewidth]{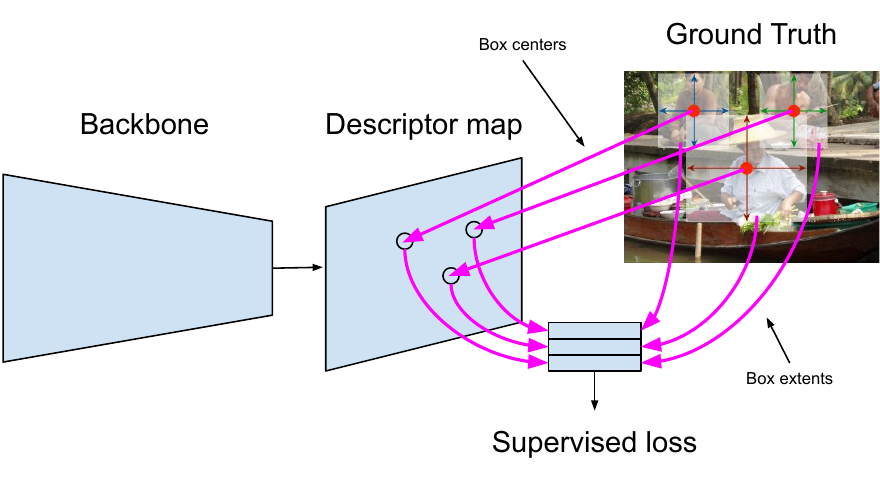}
    \caption{CenterNet supervision.  The backbone network analyzes an image and produces a dense descriptor map for each location in the image. Locations corresponding to ground truth object centers are used to predict bounding boxes and the resulting boxes are supervised using a standard bounding box regression loss.}
    \label{fig:centernet}
\end{figure}

The key idea of centernet is to represent an object as a single keypoint.
The keypoint is located at the object center, and has an associated feature vector (which I will henceforth call a descriptor) that is trained to predict properties of the object.
The network produces a dense descriptor map for every location in an image, and at training time only the descriptors that correspond to ground truth detections are supervised.
This situation is diagrammed in Figure~\ref{fig:centernet}.

In addition to bounding boxes, the descriptors are also used to produce class specific heatmaps for the different categories of objects that the detector recognizes.
At test time the model searches for peaks in the class specific heatmaps to choose detection centers, and returns bounding boxes regressed from the descriptors at the corresponding locations.
The heatmaps are class specific, and are therefore tied to the original label set used to train the model.
However, the model uses a single set of weights for bounding box prediction across all classes, making the bounding box predictions class agnostic.

\subsection{Feature based tracking}
\label{sec:autotrack}

We work in continuous normalized image coordinates.
Normalized means that the domain $[0, 1] \times [0, 1]$ covers all valid locations in the image.
Continuous means that we treat pixel data as regularly sampled grid of measurements from an underlying continuous signal where the values are sampled at the center of the pixel.
The ostensible continuous signal is recovered from pixel values using bilinear interpolation for interior values, and constant extrapolation at the boundaries.
We apply this logic both to pixel values and to the dense descriptor maps produced by centernet.

This is a convenient coordinate system to work in, if we have an image $I$ with a dense descriptor map $\phi(I)$, then the the image at location $p$ has value $I[p]$ and the corresponding descriptor is $\phi(I)[p]$, regardless of the relative sizes of the image and descriptor maps.

Suppose we have two images $I_t$ and $I_{t'}$ with $t'-t > 1$ and corresponding point annotations $p_t$ and $p_{t'}$.  From this we obtain two annotated descriptors $d_t = \phi(I_t)[p_t]$ and $d_{t'} = \phi(I_{t'})[p_{t'}]$ by sampling the dense descriptor maps at each annotated point.
Using the annotated descriptrs we can predict a descriptor track for intermediate frames $t < \tau < t'$ as
\begin{align*}
    \hat{d}_\tau = (1-\frac{\tau - t}{t' - t})d_t + \frac{\tau - t}{t' - t}d_{t'}
\end{align*}
which is just linear interpolation in descriptor space.

The interpolated descriptors are turned back into locations by computing
\begin{align*}
    \hat{p}_\tau = \arg\min_p ||\phi(I_\tau)[p] = \hat{d}_\tau||
\end{align*}
which corresponds to nearest neighbour matching of the interpolated descriptor to the corresponding image frame descriptor map.
The predicted locations can further be turned into boxes by sampling the box predictions in a similar way.

\subsection{Feature caching}

The biggest obstacle to making effective use of predictions from the centernet model during annotation is speed.
A latency of a quarter second to obtain a bounding box from a point is extremely noticeable and frustrating when you are trying to refine a selected location to get the best bounding box possible.
To make the UI work at interactive speeds we pre-process every frame in each video using the centernet backbone and cache the descriptor maps as well as the bounding box predictions for each point in each frame.
Caching features allows us to avoid re-runing the network to get a new box prediction, and makes the experience substantially smoother for the operator.

\section{Discussion}

Using the tool described in this report we can reduce the time to annotate a 300 frame video of a single target object by at least a factor of 5.3 compared to annotating each frame using extreme clicking.
This turns a 30-40 minute annotation task into one that can be completed in less than 10 minutes.
Although the labels obtained in this way appear to be lower quality they do not appear to be substantially so.
Annotating every frame with assistance from the objectness prior of centernet increases the annotation quality over tracks produced by the autotrack feature at the cost of dramatically increased annotation time (although still faster than extreme clicking). 
It remains to be seen if the quality is sufficient to train detectors to re-identifiy the target object in new videos.

UI latency was the biggest challenge in making the centernet predictions useful during annotation.
Refining a predicted bounding box by refining the selected descriptor is extremely annoying when there is noticeable latency between clicking and seeing the result.
Latency also reduces the number of refinements that can be tried before it would have been faster to draw the box directly.

In the current implementation the autotracking feature still takes 0.5-1s to generate a new track for each frame, and this latency limits the usefulness of the sparklines and the smartjump feature (since smartjump relies on the track to function).

The centernet model we used is trained on MSCOCO which has people and tables as categories.
I noticed anecdotally that the model sometimes gets distracted by these objects e.g.\ when the target novel object is near the center of a table the model will prefer to identify the table.
I observed similar behavior with a person in frame manipulating the target object as well.
This could likely remedied by removing people and tables from the original dataset and retraining the centernet (or perhaps adding a separate head that is specialized to these classes).

The extreme clicking UI would beenfit from a large crosshair cursor to help distinguish near ties for extreme points.
The authors of the extreme clicking paper mention that this is not useful, but I think it would have been helpful for annotating the pliers especially, where it was often necessairy to make judgements about the relative positions of highly separated points.
This feature is easy to implement so I see no reason not to include it in the future.

It is harder than you might expect to flim an an object from many angles with a handheld camera while keeping it consistently in frame.
The tool could benefit from the ability to exclude frames from the video so that frames can be excluded from annotation (and later from the training set for the downstream detector).
I actually implemented a basic exclusion feature before running the experiments in this report, but I found it too awkward to use, so further thought is required on how to present it.

A variant I would like to try in the future is using extreme clicking to annotate boxes and selecting descriptors to match the operator provided boxes instead of having the operator pick descriptor locations directly.
This would slow down selecting easy cases a bit but would likely speed up tricky annotation cases, and would allow the tool to identify when there is no good descriptor match for the desired annotation.

\bibliographystyle{plainnat}
\setlength{\bibsep}{5pt} 
\setlength{\bibhang}{0pt}
\bibliography{refs}

\begin{thebibliography}{3}
\providecommand{\natexlab}[1]{#1}
\providecommand{\url}[1]{\texttt{#1}}
\expandafter\ifx\csname urlstyle\endcsname\relax
  \providecommand{\doi}[1]{doi: #1}\else
  \providecommand{\doi}{doi: \begingroup \urlstyle{rm}\Url}\fi

\bibitem[Papadopoulos et~al.(2017)Papadopoulos, Uijlings, Keller, and
  Ferrari]{papadopoulos2017extreme}
Dim~P Papadopoulos, Jasper~RR Uijlings, Frank Keller, and Vittorio Ferrari.
\newblock Extreme clicking for efficient object annotation.
\newblock In \emph{Proceedings of the IEEE international conference on computer
  vision}, pages 4930--4939, 2017.

\bibitem[Zhou et~al.(2019)Zhou, Wang, and Kr{\"a}henb{\"u}hl]{zhou2019objects}
Xingyi Zhou, Dequan Wang, and Philipp Kr{\"a}henb{\"u}hl.
\newblock Objects as points.
\newblock \emph{arXiv preprint arXiv:1904.07850}, 2019.

\bibitem[Zhou et~al.(2020)Zhou, Koltun, and
  Kr{\"a}henb{\"u}hl]{zhou2020tracking}
Xingyi Zhou, Vladlen Koltun, and Philipp Kr{\"a}henb{\"u}hl.
\newblock Tracking objects as points.
\newblock \emph{arXiv preprint arXiv:2004.01177}, 2020.

\end{thebibliography}

\clearpage

\appendix

\section{User interface controls}

The application controls are largely the same across different annotation styles, although some controls are only available when the corresponding feature is enabled (e.g.\ timeline navigation requires the timeline be enabled).

Regardless of the annotation style left clicking in the viewport always selects the corresponding point, and middle clicking anywhere in the viewport clears any annotation (or partial annotation) in the current frame.

When the timeline is active frames can be selected with the mouse by clicking on the timeline.
The current frame cursor above the timeline indicates the position of the current frame in the video, and interacting with the timline (clicking, dragging, etc) behaves as you would expect a video scrubber to beahve.

There are several keyboard actions to support navigation and annotation, which are shown in the following table:

\begin{table}[h]
\centering
\begin{tabular}{ll|ll}
\textbf{T} & \textbf{A} & \textbf{Key} & \textbf{Behavior} \\ \hline
\checkmark & & \texttt{left} & Shift backwards by 1 frame \\
\checkmark & & \texttt{right} & Shift  by 1 frame \\
\checkmark & & \texttt{shift+left} & Shift backwards by 10 frames \\
\checkmark & & \texttt{shift+right} & Shift forwards by 10 frames \\
\checkmark & & \texttt{ctrl+left} & Shift backwards to the nearest annotated frame \\
\checkmark & & \texttt{ctrl+right} & Shift forwards to the nearest annotated frame \\
 & & \texttt{f} & Jump to an unannotated frame \\
\checkmark & \checkmark & \texttt{r} & Refresh the track
\end{tabular}
\end{table}

The \textbf{T} and \textbf{A} columns indicate if the timeline or autotrack must be enabled for the key to be active.
Requirements are inclusive, so refreshing the track is only possible when both the timeline and autotrack to be enabled.
The behavior of the jump key changes from random jumping to smart jumping when smartjump is enabled.





\end{document}